\documentclass{bvm} %

\DeclareMathOperator*{\argmin}{arg\,min}
\usepackage{booktabs}

\begin{document}
\newcommand{\bvmyear}{2025}

\selectlanguage{english} %

\title{iRBSM: A Deep Implicit 3D Breast Shape Model}
\author{
	Maximilian \lname{Weiherer} \inst{1,2}, 
    Antonia \lname{von Riedheim} \inst{3}, 
    Vanessa \lname{Brébant} \inst{3}, 
    Bernhard \lname{Egger} \inst{1},
	Christoph \lname{Palm} \inst{2} 
}

\authorrunning{Weiherer et al.}

\institute{
\inst{1} Visual Computing Erlangen, Friedrich-Alexander-Universtität Erlangen-Nürnberg\\
\inst{2} Regensburg Medical Image Computing (ReMIC), OTH Regensburg\\
\inst{3} Department for Plastic, Hand and Reconstructive Surgery, University Hospital Regensburg\\
}

\email{maximilian.weiherer@fau.de}

\maketitle

\begin{abstract}
We present the first deep~\textit{implicit} 3D shape model of the female breast, building upon and improving the recently proposed Regensburg Breast Shape Model (RBSM).
Compared to its PCA-based predecessor, our model employs implicit neural representations; hence, it can be trained on raw 3D breast scans and eliminates the need for computationally demanding non-rigid registration -- a task that is particularly difficult for feature-less breast shapes. 
The resulting model, dubbed iRBSM, captures detailed surface geometry including fine structures such as nipples and belly buttons, is highly expressive, and outperforms the RBSM on different surface reconstruction tasks.
Finally, leveraging the iRBSM, we present a prototype application to 3D reconstruct breast shapes from just a single image.
Model and code publicly available at \url{https://rbsm.re-mic.de/implicit.}
\end{abstract}

\section{Introduction}
\begin{figure}
    \includegraphics[width=\linewidth]{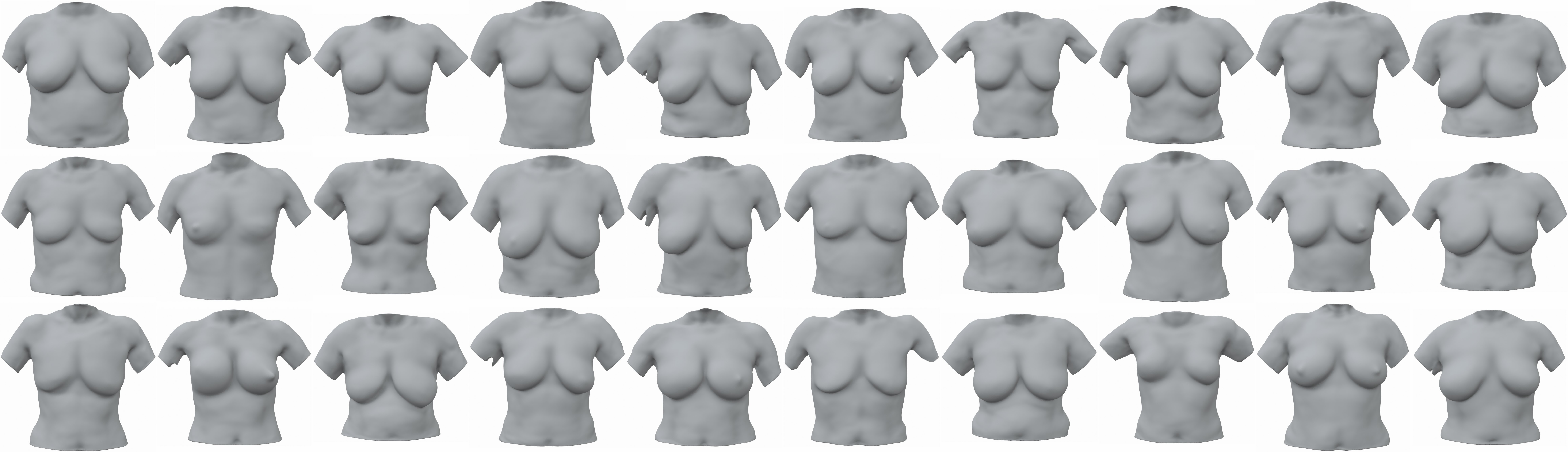}
    \caption{\textbf{Random samples from the proposed iRBSM}. Our model is able to generate a large variety of plausible breast shapes, including fine details such as the nipples or belly buttons.}
    \label{fig:random_samples}
\end{figure}

Albeit emerging use cases, 3D generative shape models for the female breast (such as the one proposed in this work, see Fig.~\ref{fig:random_samples}) rarely exist.
Most probably, this is due to the sensitivity of the data, making it challenging to acquire even small-sized datasets.
Potential applications of 3D breast shape models include breast surgery planning and simulation, breast volume estimation from 3D scans, and post-operative outcome assessment and monitoring.
Besides clinical applications, such models may also be of great interest to the fashion industry, enabling virtual try-on or custom bra design.
To date, only a single publicly available shape model of the female breasts exists: the Regensburg Breast Shape Model (RBSM)~\cite{weiherer2023}.
The RBSM is a Principal Component Analysis (PCA)-based shape model and has been built from over a hundred 3D breast scans acquired in a standing position.
While it represents an important step toward an expressive and large-scale breast model, it is not without limitations.
\begin{SCfigure}[5][t]
	\includegraphics[width=0.17\textwidth]{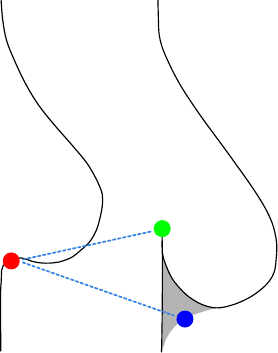}
  \caption{\textbf{The RBSM suffers from erroneous correspondences as a result of occlusions}. As shown, for large and sagged breasts, occluded underbusts (grey region) yield incorrect correspondences when using template-based surface registration. Specifically, while the red and the green points are supposed to correspond semantically, due to the occlusion of the underbust in~\textit{surface-only} 3D breast scans, the red and the blue points will incorrectly match instead. This leads to distortions on the shape boundaries in the resulting shape model.}
\label{fig:correspondence_errors}
\end{SCfigure}
First, it assumes semantically accurate point-to-point correspondence among registered 3D breast scans -- an assumption that, in fact, can not be properly satisfied on~\textit{feature-less} breast shapes using standard surface registration techniques as used in~\cite{weiherer2023}.
The problem is amplified by severe occlusions in 3D breast scans, see Fig.~\ref{fig:correspondence_errors}.
Secondly, the RBSM lacks detail, often failing to capture fine structures such as nipples. 
This is a known problem in PCA-based models and results from the linearity of dimensional reduction.
Finally, the proposed pipeline and mesh-based nature of the model itself lacks practicability,~\textit{always} requiring computationally expensive and error-prone non-rigid surface registration, including manual landmarking.

To address the aforementioned shortcomings, we introduce an~\textit{implicit} variant of the RBSM, dubbed iRBSM, which employs implicit neural representations.
As opposed to the mesh-based representation of the RBSM, we model surfaces as the zero-level set of a~\textit{Signed Distance Function} (SDF) which is parametrized by a latent-conditioned neural network.
This representation has several key benefits over linear, PCA-based shape models. 
First, no fixed topology is required, enabling implicit models to naturally handle occlusions caused by sagged breasts without producing unwanted effects such as correspondence errors. 
Notably, this also renders surface registration (at all stages -- during model building and at inference) unnecessary. 
Secondly, SDF-based generative models allow for a higher level of detail due to their non-linear nature. 
The result is an expressive model that can generate highly realistic and diverse breast shapes, see Fig.~\ref{fig:random_samples}.

\textbf{Related work. }
In 2007, Seo et al.~\cite{seo2007} were the first to build a classical, PCA-based 3D shape model from 28 breast scans to analyze breast volume and surface measurements.
They assumed symmetric breasts.
Recently, Mezier et al.~\cite{mazier2021} proposed a~\textit{rigged} breast model built from 55 artist-created blendshapes.
Due to the synthetic nature of the data, their model generates rather unnatural-looking breast shapes.

\section{Materials and Methods}
We will now describe the iRBSM in detail, starting with the training data that has been used to build our model, followed by the model formulation and implementation details.

\subsection{3D Breast Scan Database}
Our model is trained on the same dataset already used for the RBSM but extended by 58 scans, totaling 168 consistently oriented scans.
All scans have been taken in a standing position using the portable Vectra H2 system (Canfield Scientific, New Jersey, USA) which employs photogrammetry to reconstruct 3D surfaces from three photographs taken from frontal and lateral views.
Data has been collected between 2020 and 2023 at the St. Josef Hospital Regensburg. 
Please see~\cite{weiherer2023} for more information.

\textbf{Data pre-processing. }
Our raw 3D breast scans are not closed, watertight surfaces -- a necessary condition to be representable as SDF.
To close the raw input scans, we use the following two-stage workflow. 
First, we offset the surface mesh in negative $z$-direction (\textit{i.e.}, we copy the original mesh and translate it in negative $z$-direction) and connect corresponding border vertices.
Next, we cut the mesh along (i) the frontal plane passing through the lateral breast pole (see~\cite{hartmann2020}) and (ii), the transversal plane passing through a point approximately 2 cm below the belly button. 
After each step, we again connect border vertices, eventually leading to a closed and watertight surface mesh. 

\subsection{Model Formulation}
\label{subsec:model_formulation}
We represent the iRBSM as a function $\phi:\mathbb{R}^3\times\mathbb{R}^L\rightarrow\mathbb{R}$, mapping a 3-dimensional coordinate $x$ and an $L$-dimensional latent code $z$ to a signed distance, $\phi(x,z)=\Phi_\theta([x, z])$, where $\Phi_\theta:\mathbb{R}^{3+L}\rightarrow\mathbb{R}$ is a~\textit{latent-conditioned} SDF parametrized by a neural network with weights $\theta$.
Once $\phi$ is trained, the surface $\mathcal{S}_z\subset\mathbb{R}^3$ of a shape corresponding to the latent code $z$ is given as the zero-level set of $\phi$, that is, $\mathcal{S}_z=\{x\in\mathbb{R}^3\mid\phi(x,z)=0\}$. 
The surface $\mathcal{S}_z$ can be readily extracted as triangular mesh using marching cubes algorithm.

\textbf{Training. }
The auto-decoder framework~\cite{park2019} is used for training, in which we simultaneously optimize the network's parameters $\theta$ and latent codes $z$.
Given a dataset of 3D breast scans each of which represented as an oriented point cloud, $X=\{x_1,x_2,\dots,x_n\}\subset\mathbb{R}^3$ and associated per-point normals $N=\{n_1,n_2,\dots,n_n\}\subset\mathbb{R}^3$, we employ the following loss proposed in~\cite{gropp2020} with slight modification:
\begin{multline}
    \mathcal{L}(\theta,z)=\sum_{i=1}^n|\phi(x_i,z)|+\Vert\nabla_{x_i}\phi(x_i,z)-n_i\Vert_2+\\
    \lambda_1\mathbb{E}_{x\sim D}\left[|\Vert\nabla_x\phi(x,z)\Vert_2-1|+\exp(-\alpha|\phi(x,z)|)\right]+\lambda_2\Vert z\Vert_2,
    \label{eq:loss}
\end{multline}
where $D:=[-1,1]^3\cup\{x_1+\epsilon_1,x_2+\epsilon_2,\dots,x_n+\epsilon_n\}$ with $\epsilon_i\sim\mathcal{N}(0,\sigma^2)$ is a set of free-space points.
Please note that we do \textit{not} require ground truth SDF values for supervision; the whole model is trained solely on the raw 3D breast scans.

\textbf{Inference. }
Given a 3D breast scan as \textit{unoriented} point cloud, $X'=\{x'_1,x'_2,\dots,x'_m\}$, at test-time, we obtain the corresponding latent code $z^*$ by fixing the model's parameters $\theta$ and optimizing
\begin{equation}
    z^*=\argmin_{z\in\mathbb{R}^L}\left\{\sum_{i=1}^m|\phi(x'_i,z)|+\lambda\Vert z\Vert_2\right\}.
    \label{eq:latent_optim}
\end{equation}
Here, $\lambda\geq 0$ can be used to filter out noise.
We refer to~\cite{park2019} for further details.

\subsection{Implementation Details}
We represent $\phi$ as an 8-layer fully-connected MLP with 512-dimensional hidden layers and skip connection to the middle layer.
The final SDF value is regressed by applying the softplus activation function.
Moreover, we use the geometric initialization scheme proposed in~\cite{gropp2020} and a latent dimension of 256.
Latent codes are initialized from a zero-mean normal distribution with variance $10^{-4}$.
We trained for 10,000 epochs using a batch size of 16 and the Adam optimizer with a learning rate of $5\times 10^{-4}$ for model parameters and $10^{-3}$ for latent codes. 
Both learning rates are decayed by a factor of 0.5 every 2,000 epochs.
We use 5,000 on-surface points and 5,000 free-space points sampled from $D$ for each shape.
We empirically set $\lambda_1=0.1,\lambda_2=0.01$, and $\alpha=10$.
All experiments are run on a single NVIDIA RTX A5000 with 24 GB of VRAM. 
Training the iRBSM took about 20 hours.
At inference, we optimize Eq. (\ref{eq:latent_optim}) for 5,000 iterations using Adam.
Additionally, we decay the learning rate by a factor of 0.5 every 1,000 iterations.
We set $\lambda=0.01$ in the noise-free case, and use $\lambda=0.1$ for noisy inputs.

\section{Results}
Our evaluation is based on a test set of ten 3D breast scans held out during training.

\begin{table}
\begin{tabular*}{\textwidth}{l@{\extracolsep\fill}cccccc}
\toprule
& \multicolumn{3}{c}{\textbf{Sparse} (1,000 points)} & \multicolumn{3}{c}{\textbf{Dense} ($\approx$ 20,000 points)} \\
\cmidrule(lr){2-4} \cmidrule(lr){5-7}
& CD $\downarrow$ & F-Score $\uparrow$ & NC $\uparrow$ & CD $\downarrow$ & F-Score $\uparrow$ & NC $\uparrow$ \\
\midrule
RBSM~\cite{weiherer2023} & 4.35 {\tiny $\pm$ 1.27} & 43.3 {\tiny $\pm$ 16.38} & 96.1 {\tiny $\pm$ 1.89} & 4.27 {\tiny $\pm$ 1.24} & 44.9 {\tiny $\pm$ 16.73} & 96.1 {\tiny $\pm$ 1.87} \\
iRBSM (Ours) & \textbf{1.66} {\tiny $\pm$ 0.29} & \textbf{88.4} {\tiny $\pm$ 6.54} & \textbf{98.6} {\tiny $\pm$ 0.83} & \textbf{1.57} {\tiny $\pm$ 0.28} & \textbf{90.7} {\tiny $\pm$ 6.06} & \textbf{98.8} {\tiny $\pm$ 0.75} \\
\bottomrule
\end{tabular*}
\caption{\textbf{Quantitative results for sparse surface reconstruction}. Our model consistently outperforms the RBSM by a large margin. For comparison, we also report results on dense point clouds.}
\label{tab:quant_results_recon}
\end{table}

\begin{figure}[t]
    \includegraphics[width=\linewidth]{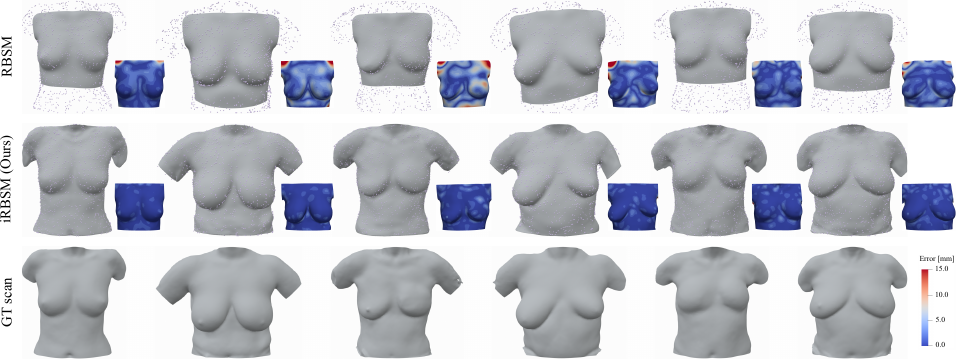}
    \caption{\textbf{Qualitative results for sparse surface reconstruction}. We show surface reconstructions obtained from fitting the RBSM and our model to sparse point clouds. Our model leads to visually more pleasant reconstructions than its PCA-based counterpart.}
    \label{fig:qual_results_recon}
\end{figure}

\textbf{Sparse surface reconstruction. }
We start by evaluating our model's ability to reconstruct accurate surfaces from sparse point clouds. 
To do so, we randomly sample 1,000 surface points from each of our test scans to which we then fit our model, see Section~\ref{subsec:model_formulation}.
We compare against the RBSM\footnote{We took the augmented version including mirrored scans and use all 219 available principal components.}~\cite{weiherer2023}, which we fit to point clouds using landmark-based rigid alignment followed by model-based non-rigid surface registration. 
For the latter, we optimize a cost function consisting of a distance, landmark, and regularization term using Adam.
We manually clicked five landmarks.

We report Chamfer distance (CD), F-Score (with a threshold of 2.5 mm), and normal consistency (NC).
To ensure fair comparison, we compute metrics only within the breast region, defined by cropping meshes using axis-aligned planes passing through the upper breast poles, along axillary lines, and approximately 5 cm below the lower breast poles.

Results can be found in Tab.~\ref{tab:quant_results_recon} and Fig.~\ref{fig:qual_results_recon}.
Our model quantitatively outperforms the RBSM by a large margin (over 2.5 times better in CD) while also leading to visually more pleasant results, overall being much closer to the ground truth (GT) than surface reconstructions obtained from the RBSM.
Moreover, fitting the iRBSM took approximately 20 seconds on average and did \textit{not} require any pre-processing; in contrast, fitting the RBSM required more than 23 seconds, \textit{excluding} manual landmarking.

\begin{figure}[t]
   \includegraphics[width=\linewidth]{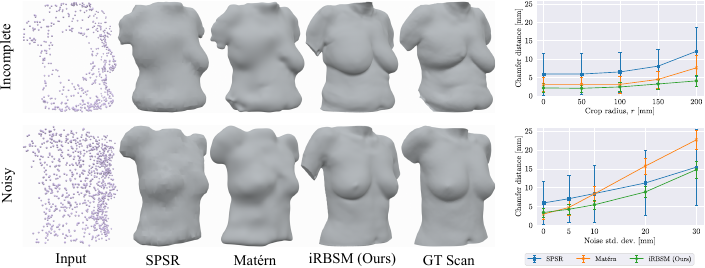}
    \caption{\textbf{Qualitative and quantitative results for incomplete (top) and noisy (bottom) inputs}. In both scenarios, our model leads to visually more appealing yet plausible reconstructions. This is also reflected quantitatively (right) when compared to learning-free methods, SPSR~\cite{kazhdan2013} and Matérn~\cite{weiherer2024}. Qualitative results are shown for $r=100$ (top) and $\sigma=10$ (bottom).}
    \label{fig:results_inc_noisy}
\end{figure}
 
\textbf{Surface reconstruction from incomplete and noisy scans. }
Next, we evaluate the iRBSM in a more realistic setting where only incomplete and noisy scans are available.
To simulate holes in scans, we again randomly sample 1,000 surface points from each scan of our test set and then cut away a random patch by removing points within a sphere with a pre-defined crop radius, $r\geq 0$, centered on an arbitrary surface point.
Noise is simulated by adding Gaussian noise with various standard deviations, $\sigma\geq 0$, to surface points. 
We compare against model-free approaches, Screened Poisson Surface Reconstruction (SPSR)~\cite{kazhdan2013}, and a recent kernel-based reconstruction approach~\cite{weiherer2024}.
For SPSR, we did a modest parameter sweep, considering all possible combinations of the~\textit{octree depth} in $\{6,7,8,9\}$ and \textit{point weight} in $\{4,100,1000\}$.
For~\cite{weiherer2024}, we varied the regularization weight $\lambda\in\{0,10^{-13},10^{-12},10^{-11},10^{-10}\}$ and bandwidth $h\in\{0.5,1,2\}$.

Qualitative and quantitative results can be found in Fig.~\ref{fig:results_inc_noisy}, demonstrating that the iRBSM is, among the methods tested, the best to handle large holes and even severe noise.
This demonstrates that the iRBSM is indeed a strong prior for 3D surface reconstruction of real-world point cloud data -- a quite practical finding that we will exploit next.

\begin{figure}[t]
    \centering
    \includegraphics[width=\linewidth]{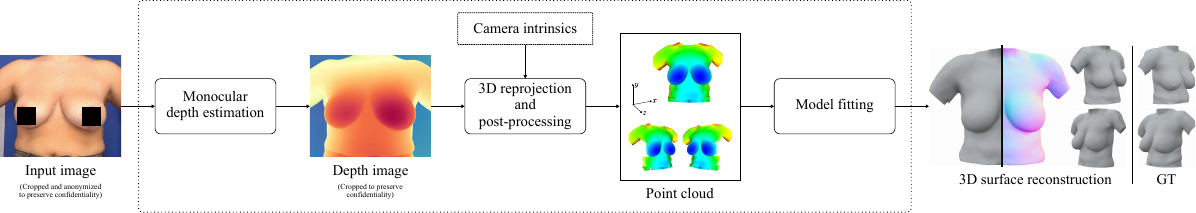}
    \caption{\textbf{Single image to 3D using the iRBSM}. We employ our model to 3D reconstruct breast shapes from \textit{single} images by first estimating depth using a state-of-the-art monocular depth estimation model~\cite{yang2024} and then fitting the iRBSM to the unprojected depth map. This yields high-quality surface reconstructions with accurate normals in seconds and without special hardware.}
    \label{fig:3d_from_2d_application}
\end{figure}

\textbf{Sample application: single image to 3D. }
Lastly, we prototype an application in which we use the iRBSM to 3D reconstruct breast shapes from a~\textit{single} image, see Fig.~\ref{fig:3d_from_2d_application}.
Given an image from a calibrated camera, we employ a state-of-the-art monocular depth estimation model, Depth Anything V2~\cite{yang2024}, reproject the obtained depth map into 3D leveraging pre-computed camera intrinsics, and finally fit our model to the resulting point cloud.
As shown, this workflow is capable of reconstructing high-quality surfaces.

\section{Conclusion}
The proposed iRBSM presents a significant advancement toward an expressive 3D breast model, being able to recover plausible breast shapes from sparse, incomplete, and noisy observations in few seconds.
Moreover, we believe the proposed application has the potential to become a practical and robust method to acquire accurate 3D breast reconstructions from images without expensive hard- and software or special setups.

\printbibliography

\end{document}